\documentclass[10pt, conference, letterpaper]{IEEEtran}
\usepackage{}
\usepackage{latexsym,bm,amsmath,amssymb}
\usepackage{cite}
\usepackage[ruled]{algorithm2e}
\usepackage{graphicx}
\usepackage{subcaption}
\usepackage{color}

\usepackage[flushleft]{threeparttable}
\usepackage{balance}
\usepackage{array}
\usepackage{multirow}

\newcommand{\alg}{mGAN-AI}
\newcolumntype{C}[1]{>{\centering\let\newline\\\arraybackslash\hspace{0pt}}m{#1}}

\begin{document}

\title{
Beyond Inferring Class Representatives: User-Level Privacy Leakage From Federated Learning
}


\author{\IEEEauthorblockN{Zhibo Wang$^{\dagger}$, Mengkai Song$^{\dagger}$, Zhifei Zhang$^{\ddagger}$, Yang Song$^{\ddagger}$, Qian Wang$^{\dagger}$, Hairong Qi$^{\ddagger}$}
\IEEEauthorblockA{$^{\dagger}$School of Cyber Science and Engineering, Wuhan University, P. R. China}
\IEEEauthorblockA{$^{\ddagger}$Deptment of Electrical Engineering and Computer Science, University of Tennessee, Knoxville, USA}
Email: \{zbwang, mksong, qianwang\}@whu.edu.cn, \{zzhang61, ysong18, hqi\}@utk.edu
}

\maketitle
\IEEEpeerreviewmaketitle

\begin{abstract}
Federated learning, i.e., a mobile edge computing framework for deep learning, is a recent advance in privacy-preserving machine learning, where the model is trained in a decentralized manner by the clients, i.e., data curators, preventing the server from directly accessing those private data from the clients. This learning mechanism significantly challenges the attack from the server side. Although the state-of-the-art attacking techniques that incorporated the advance of Generative adversarial networks (GANs) could construct class representatives of the global data distribution among all clients, it is still challenging to distinguishably attack a specific client (i.e., user-level privacy leakage), which is a stronger privacy threat to precisely recover the private data from a specific client. This paper gives the first attempt to explore user-level privacy leakage against the federated learning by the attack from a malicious server. We propose a framework incorporating GAN with a multi-task discriminator, which simultaneously discriminates category, reality, and client identity of input samples. The novel discrimination on client identity enables the generator to recover user specified private data. Unlike existing works that tend to interfere the training process of the federated learning, the proposed method works ``invisibly'' on the server side. The experimental results demonstrate the effectiveness of the proposed attacking approach and the superior to the state-of-the-art.\footnote{This manuscript has been accepted by IEEE INFOCOM 2019.}
\end{abstract}

\section{Introduction}

\begin{figure}[t]
	\centering
	\includegraphics[width=0.95\columnwidth]{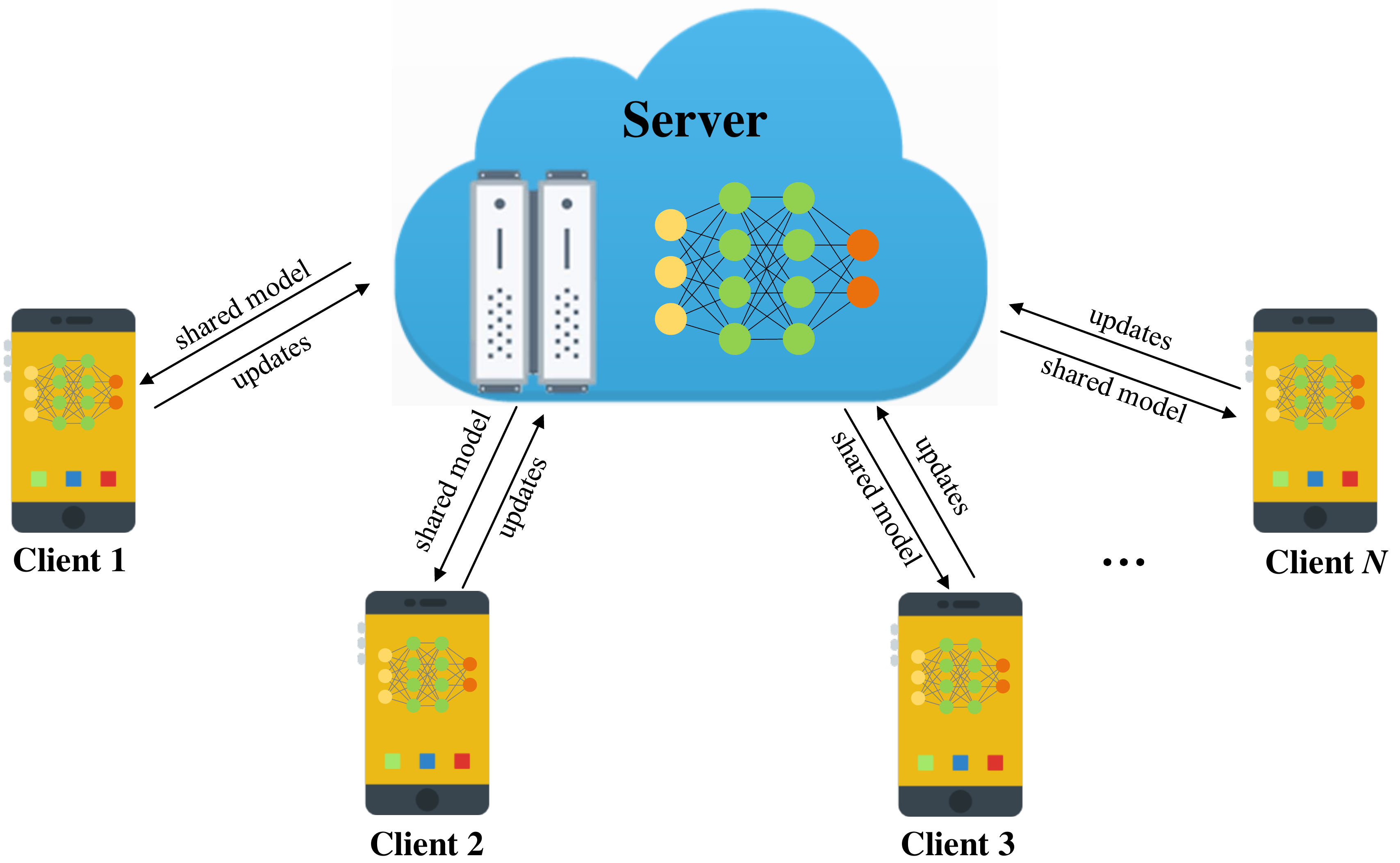}
	\caption{The framework of federated learning. The server sends the shared model to each client, which trains the shared model locally by its private data. Then, the update from each client is uploaded to the server, where those updates are aggregated/averaged to improve the shared model in a collaborative manor.}
	\label{fig:fedLearning}
	\vspace{-3mm}
\end{figure}

Increasingly, modern deep learning technique is beginning to emerge in the networking domain, such as a crowdsourced system for learning tasks.
But utilizing conventional centralized training methodology requires local storage of the crowdsourced data, which suffers from the high volume of traffic, highly computational demands and privacy concerns. 
For collectively reaping the benefits of the shared model trained from this rich data without the need to store it centrally, distributed learning framework was proposed, serving as a mobile edge computing framework for deep learning.
Shokri et al.~\cite{shokri2015privacy} proposed the collaborative learning DSSGD, where the data providers, i.e., clients, train locally on a shared model, and then the server will collect those local models/updates to estimate/update a global model instead of directly assessing the private data from clients. Then, the global model is sent back to clients, iterating the aforementioned local training process.
In the same token, federated learning~\cite{mcmahan2016communication} proposed a variant of decentralized learning with higher efficiency. The key improvement lies in the way of updating the global model, i.e., DSSGD performs the aggregated update while the federated learning conducts the averaged update, which is more suitable for the commonly non-IID and unbalanced data distribution among clients in the real world. Fig.~\ref{fig:fedLearning} illustrates the framework of federated learning.

However, recent works demonstrated that the collaborative learning is vulnerable to the inference attack, e.g., reconstruction attack and membership attack, by malicious servers/clients because the shared model is updated according to those private data, whose pattern is encoded into the model parameters. Therefore, if a corresponding decoder could be constructed, the private data or statistics will be recovered inversely. Under the assumption of a malicious server, Aono et al.~\cite{aono2017privacy} partially recovered the private data based on the observation that the gradient of weights is proportional to that of the bias at the first layer of the model, and their radio approximates to the training input.
But it could only apply to a simplified setting where the training batch has only one sample.
Hitaj et al.~\cite{hitaj2017deep} proposed a GAN-based reconstruction attack against the collaborative learning by assuming a malicious client, which utilized the shared model as the discriminator to train a GAN~\cite{goodfellow2014generative}. Eventually, the learned generator (equivalent to the decoder) will successfully mimic the training samples.
Although \cite{hitaj2017deep} had demonstrated the effectiveness of GAN-based attack against DSSGD,
it presents mainly three limitations: 1) It would change the architecture of the shared model and introduce adversarial influence to the learning process, while the former unrealistically assumes a powerful malicious client, and the latter compromises the learning process. 2) The GAN-based attack suffers from performance degradation when attacks the federated learning because the adversarial influence introduced by the malicious client would become trivial after the average update. 3) In addition, the GAN-based attack could only infer class-wised representatives, which are generic samples characterizing class properties rather than the exact samples from the clients~\cite{melis2018inference}.

Motivated by those drawbacks in existing reconstruction attacking techniques,
we propose a more generic, practical, and invisible attacking framework against the federated learning without affecting the learning process. In addition, we further target client-level privacy\footnote{In this paper, ``user-level'' and ``client-level'' are used interchangeably.},
which is more challenging and empirically meaningful than globally recovering the data representatives that characterize the population distribution as in the GAN-based attack.
Instead of assuming a malicious client, the proposed method performs on the server side, which first explores the attack by a malicious server. Specifically, we adapt the original GAN to a multi-task scenario where reality, category, and identity of the target client are considered simultaneously.
On the one hand, performing additional tasks when training the GAN improves the quality of synthesized samples~\cite{odena2016conditional} without the necessity of modification on the shared model or compromising to the federated learning, achieving invisible attack. On the other hand, the novel task of discriminating client identity enables client-level privacy recovery.
For simplicity, the proposed framework is referred to as multi-task GAN for Auxiliary Identification (\alg). 

The key novelty of
\alg~is to compromise the client-level
privacy
that is defined as the client-dependent properties, which specifically characterize the corresponding client and distinguish the client from the others. The key of achieving the discrimination on client identity is to obtain data representatives from each individual client, such that the training of GAN can be supervised by the identity representatives, generating samples with specific identity (i.e., from a specific client). Since the client data is unaccessible, we adopt an optimization-based method to estimate such data representatives from those accessible updates from the clients. The \alg~relaxes the assumption that clients own mutually exclusive class labels, achieving a more generic attack to the federated learning.

In summary, our contributions are mainly in three-fold:
\begin{itemize}

\item To the best of our knowledges, we first explore the attack from the perspective of a malicious server against the federated learning. In addition, beyond inferring class-wised representatives, we step further to recover user-level privacy in an invisible manner.

\item Correspondingly, we propose the generic attacking framework \alg~that incorporates a multi-task GAN, which conducts a novel discrimination on client identity, achieving attack to user-level privacy.

\item Exhaustive experimental evaluation is conducted to demonstrate the effectiveness and superior of the proposed \alg~as compared to existing works. On the MNIST and AT\&T datasets, \alg~successfully recovers the samples from a specific user.

\end{itemize}

The rest of this paper is organized as follows. Section~\ref{sec:related} introduces the related works on privacy-preserving distributed learning and attack models. Section~\ref{sec:preliminary} briefs the background knowledge. The threat models are discussed in section~\ref{sec:threat}, and then the proposed \alg~attack is detailed in sections~\ref{sec:attack}. Extensive experimental evaluation is conducted in section~\ref{sec:experiment}. Finally, section~\ref{sec:conclusion} concludes the paper.

\section{Related work}\label{sec:related}

This section provides the background of privacy-preserving related learning methods, as well as corresponding attacking approaches.

\subsection{Privacy-preserving Distributed Learning}

Distributed learning frameworks can be categorized according to the updating mechanism of the shared model. \cite{shokri2015privacy} summarized the updates while~\cite{mcmahan2016communication,konevcny2016federated}
averaged the updates, which outperform asynchronous approaches in aspects of communication efficiency and privacy preservation~\cite{chen2016revisiting}. Existing works on privacy-preserving distributed learning mostly utilize either differential privacy mechanism (DP) or secure multi-party computation (MPC).
Pathak et al.~\cite{pathak2010multiparty} proposed a DP-based global classifier by aggregating the locally trained classifiers. Hamm et al.~\cite{hamm2016learning} leveraged knowledge transfer from ensemble local models to a global differentially private model. DSSGD~\cite{shokri2015privacy} was the first collaborative learning mechanism dealing with practical distribution, which used sparse vector~\cite{dwork2014algorithmic} to realize DP.
Recently, participant-level differentially private distributed learning method~\cite{geyer2017differentially}
was proposed to further protect the membership information of participants.
Utilizing MPC, Mohassel et al.~\cite{mohassel2017secureml} presented the SecureML framework where a global model is trained on the clients' encrypted data among two non-colluding servers. Other methods based on encryption include secure sum protocol~\cite{danner2015fully} that uses a tree topology, homomorphic encryption, and secure aggregation protocol~\cite{bonawitz2017practical}.

\subsection{Attacks on the Learning Model}
\label{subsec:relate_attack}

The inference attack, e.g., membership attack and reconstruction attack, aims to infer the private features of the training data against a learning model. The membership attack~\cite{shokri2017membership,hayes2017logan,melis2018inference} is to decide whether a given sample belongs to the training dataset.
Shokri et al.~\cite{shokri2017membership} mounted the attack by utilizing the difference between model outputs from training and non-training inputs, respectively.
Hayes et al.~\cite{hayes2017logan} used GAN to detect overfitting of the target model and recognize inputs through the discriminator which could distinguish different data distributions.
Melis et al.~\cite{melis2018inference} first proposed the membership attack against the collaborative/federated learning and further inferred the properties by a batch property classifier.
By contrast, the reconstruction attack aims to construct the training samples by accessing a learning model~\cite{fredrikson2015model,hitaj2017deep,aono2017privacy}.
Fredrikson et al.~\cite{fredrikson2015model} proposed the model inversion (MI) attack to recover images from a face recognition service by maximizing the confidence values with respect to a white noise image.
Against collaborative/federated learning framework, Aono et al.~\cite{aono2017privacy} partially recovered private data of the clients based on the proportionality between the training data and the updates sent to the server (assuming a malicious server).
Hitaj et al.~\cite{hitaj2017deep} assumed a malicious client in the federated learning, where the malicious client has white-box access to the model at every training iteration. Utilizing GAN to synthesize samples, this attack successfully reconstructed the private data of other clients.

This paper proposes a GAN-based reconstruction attack against the federated learning from the perspective of a malicious server, unlike related works that failed to threaten the federated learning in a practical and invisible manner. For example, \cite{aono2017privacy} could only be applied to a simplified setting where the training batch has only one sample,
and \cite{hitaj2017deep} performed the attack by modifying the architecture of the shared model which is beyond the power of a malicious client, and it introduced negative influence into the standard training process, tending to worsen the shared model.

\section{Preliminary} \label{sec:preliminary}

\subsection{Federated Learning}
\label{sec:fedAvg}
Compared to the conventional training methods that require to directly access private data from clients, federated learning presents significant advantages in privacy preservation because of the distributed learning, where the clients share and train the model locally without uploading their private data to the server.
Fig.~\ref{fig:framework} illustrates the proposed \alg, as well as the federated learning. There are $N$ clients, each of which owns its private dataset. During the learning, clients agree on a common objective and model structure. At each iteration, the parameters of current model are downloaded from the server to clients, and then the model is trained locally on each client. Finally, those local updates are sent back to the server, where the updates are averaged and accumulated to the current shared model. Eq.~\ref{eq:federated} expresses the update of the shared model.
\begin{equation}
M_{t+1} = M_t + \frac{1}{N} \sum_{k=1}^{N}u^k_t,
\label{eq:federated}
\end{equation}
where $M_t$ denotes the shared model at the $t$th iteration, and $u^k_t$ indicates the update from the $k$th client at iteration $t$. 
\cite{mcmahan2016communication} demonstrated that a valid model could be achieved by averaging the updates only if the clients receive the model with the same initialization. The federated learning satisfies this condition.

\subsection{Generative Adversarial Networks}
\label{sec:dl}
Generative adversarial nets (GANs) were first proposed by Goodfellow~\cite{goodfellow2014generative}, which could generate samples indistinguishable from those training/real samples.
It consists of a generator $G$ and a discriminator $D$. The generator $G$ draws random samples $z$ from a prior distribution (e.g., Gaussian or uniform distribution) as the inputs, and then $G$ generates samples from $z$. Given a training set, the discriminator $D$ is trained to distinguish the generated samples from the training (real) samples. Eq.~\ref{eq:gan} shows the objective of GANs.
\begin{equation}
\begin{split}
\min_G\max_D\;V(D,G) = & \mathbb{E}_{x\sim p_{data}(x)}[\log D(x)]+\\
~ & \mathbb{E}_{z\sim p_{z}(z)}[\log (1-D(G(z)))],
\end{split}
\label{eq:gan}
\end{equation}
where $p_{data}$ and $p_{z}$ denote the training/real distribution and prior distribution, respectively. The two models $G$ and $D$ are trained alternately until this minimax game achieves Nash equilibrium, where the generated samples are difficult to be discriminated from the read ones. 
Theoretically, the global optimum is achieved at $p_{data} = p_{g}$~\cite{goodfellow2014generative}, where $p_g$ indicates the distribution of generated samples.

\section{Threat Model}
\label{sec:threat}

\begin{figure*}[!ht]
  \centering
  \includegraphics[width=1.8\columnwidth]{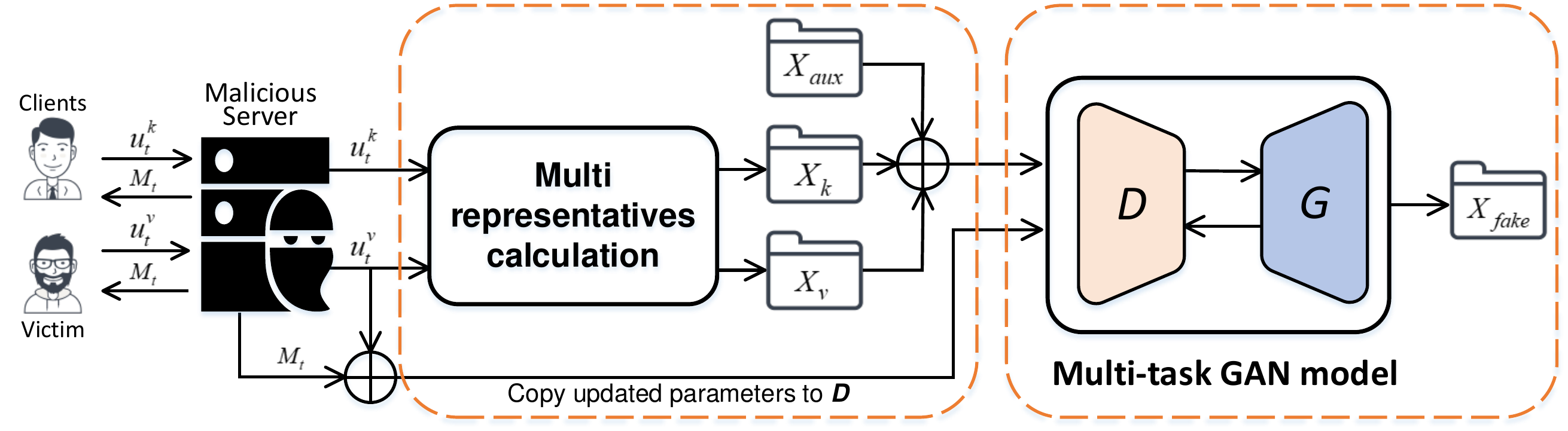}
  \caption{Illustration of the proposed \alg~from a malicious server in the federated learning. There are $N$ clients, and the $v$th client is attacked as the victim. The shared model at the $t$th iteration is denoted as $M_t$, and $u_t^k$ denotes corresponding update from the $k$th client. On the malicious server, a discriminator $D$ (orange) and generator $G$ (blue) are trained based on the update $u_t^v$ from the victim, the shared model $M_t$, and representatives $X_k$, $X_v$ from each client. $X_{aux}$ denotes an auxiliary real dataset to train $D$ on the real-fake task.}
  \label{fig:framework}
  \vspace{-4mm}
\end{figure*}

This section introduces the threat model of the proposed \alg~framework, and then the generality and rationality of assuming a malicious server in \alg~is demonstrated as compared to related works.

\subsection{Learning Scenario}
Following the training procedure of federated learning as described in Sec. \ref{sec:fedAvg}, we assume $N$ ($N \ge 2$) clients that agree on a common learning objective and train collaboratively on a shared model.
Without loss of generality, the data of all clients is considered as non-IID distributed, which is also consistent with the settings of federated learning~\cite{mcmahan2016communication}. Specifically, the data distribution of an arbitrary client $k$ ($k=1,2,\cdots,N$) is independent from that of the other clients, and thus bias from the global distribution of all clients. The biased distribution $p_k$ of the client $k$ is considered as its client-level privacy.
For simplicity, the shared model adopts an image classifier in the rest of this paper.

\subsection{Malicious Server}
The server in the federated learning is assumed to be malicious, aiming to reconstruct private data of the victim, which refers to the target client, for implementing the client-level privacy attack.
The malicious server could either only analyze the periodic updates from the clients (i.e., \textit{passive attack}), or even intentionally isolate the shared model trained by the victim to achieve more powerful attack (i.e., \textit{active attack}).
By contrast, most existing attacks against the collaborative/federated learning assumed malicious clients, which are limited at the stage of recovering class-wised representatives rather than prying client-level privacy because the malicious clients can only access aggregated updates (contributed by all the clients) from the server, while client-level attack requires individual update from each client.
Although there are attacks targeting at certain clients~\cite{hitaj2017deep,melis2018inference}, they impractical assumed that class labels of the clients are mutually different or required extra information of the target client, e.g., class labels or other client-wise properties.
More importantly, they would compromise the performance of the shared model. For example, \cite{melis2018inference} changed the training objective, and \cite{hitaj2017deep} changed the shared model and introduced mislabeled samples into the training. Considering the limitation and drawback of malicious clients, we assume a malicious server that would overcome all above problems.
The malicious server rigidly follows the rules of federated learning. Meanwhile, the passive and active attacks are performed without effecting on the learning process, i.e., shared model, objective, updates, etc.


\subsection{Communication Protocol}
In our threat model, the shared model is exchanged between the server and clients in plain text like \cite{mcmahan2016communication}, rather than in a encryption-based protocol~\cite{corrigan2013proactively,bonawitz2017practical}. Intuitively, the encryption-based aggregation prevents the adversary from accessing the updates of the clients, thus preserving their privacy.
However, recent works~\cite{hitaj2017deep,melis2018inference}, which only accessed the aggregated updates, could compromise the clients' privacy by introducing adversarial influence. In addition,
the encryption-based aggregation also prevents the server from evaluating the utility of clients' updates, degrading the learning model when malicious clients exist.
\cite{bonawitz2017practical} discussed the weakness of encryption-based protocol that it cannot guarantee the correctness of federated learning if there are malicious clients.

\section{Approach}
\label{sec:attack}

This section details the proposed \alg~attack against the federated learning for reconstructing private data of a specific victim. The main idea is to design a multi-task GAN that could discriminatingly learn the real data distribution of the victim.
Section~\ref{subsec:overview} gives a high-level overview of \alg, and Section~\ref{subsec:structure} details the structure of the GAN with a novel multi-task discriminator, which first achieves discrimination on client identity.
Then, sections~\ref{subsec:passive} and \ref{subsec:active} discuss the passive attack and active attack, respectively. The former performs in an invisible fashion, while the latter slightly interferes the share model but achieving more powerful attack. Finally, section~\ref{subsec:client} further details the inference of client-level privacy.

\subsection{Overview of \alg}
\label{subsec:overview}

The principle of GAN is to train a discriminator on target data and generated data simultaneously. Eventually, the coupled generator would yield samples close to target ones.
In the proposed \alg, the idea of training a GAN is borrowed, but the malicious server cannot access the target data (i.e., client-level private data) in the scenario of federated learning.
However, the shared model is trained locally on each client, equivalent to training a discriminator (with the same structure as the shared model) on the target data. Intuitively, directly obtaining the target data to train a discriminator is equivalent to obtaining the network update after feeding the target data to the discriminator. In the federated learning, the only accessible information to the malicious server is the updates from clients, which provide the update to the discriminator after training on the target data. For the generated data, it can be easily obtained from the corresponding generator. Therefore, the advance of GAN could be utilized in the federated learning. In addition, since GANs could generate conditioned samples if supervised by the condition in training~\cite{chen2016infogan}, we could train \alg~conditoned on the updates from victim, thus it could generate victim-conditioned samples, i.e., client-level privacy.

Fig.~\ref{fig:framework} overviews the proposed \alg. Assume $N$ clients, and the $v$th client is the victim whose data would be reconstructed by the malicious server. The malicious server acts as a normal server in the federated learning, while at the same time it is also an adversary. At the $t$th iteration, the malicious server sends the current shared model $M_t$ to each of the $N$ clients and then receives the updates $u^1_t, u^2_t,\cdots, u^{N}_t$ from them after training on their private data. Specially, $u^v_t$ denotes the update from the victim.
For reconstructing the private data of the victim, we propose a new variant of GAN with a multi-task discriminator, which simultaneously discriminates category, reality, and client identity of input samples. Noted that the structure of the discriminator is the same as the shared model regardless of the output layer.
Thus, the update $u_t^v$ from the victim $v$ could be either aggregated to the latest shared model
and then to update the discriminator
(passive attack) or aggregated to the discriminator directly (active attack).
Besides, representatives of each client are calculated to supervise the training of $D$ on identity.
Then, the updated discriminator is trained on generated samples from the generator. Correspondingly, the generator is updated to approach the data from the victim. In the following, we will introduce the structure of the proposed multi-task GAN model.

\begin{figure}[t]
    \centering
    \begin{subfigure}[b]{.48\columnwidth}
        \includegraphics[width=\textwidth]{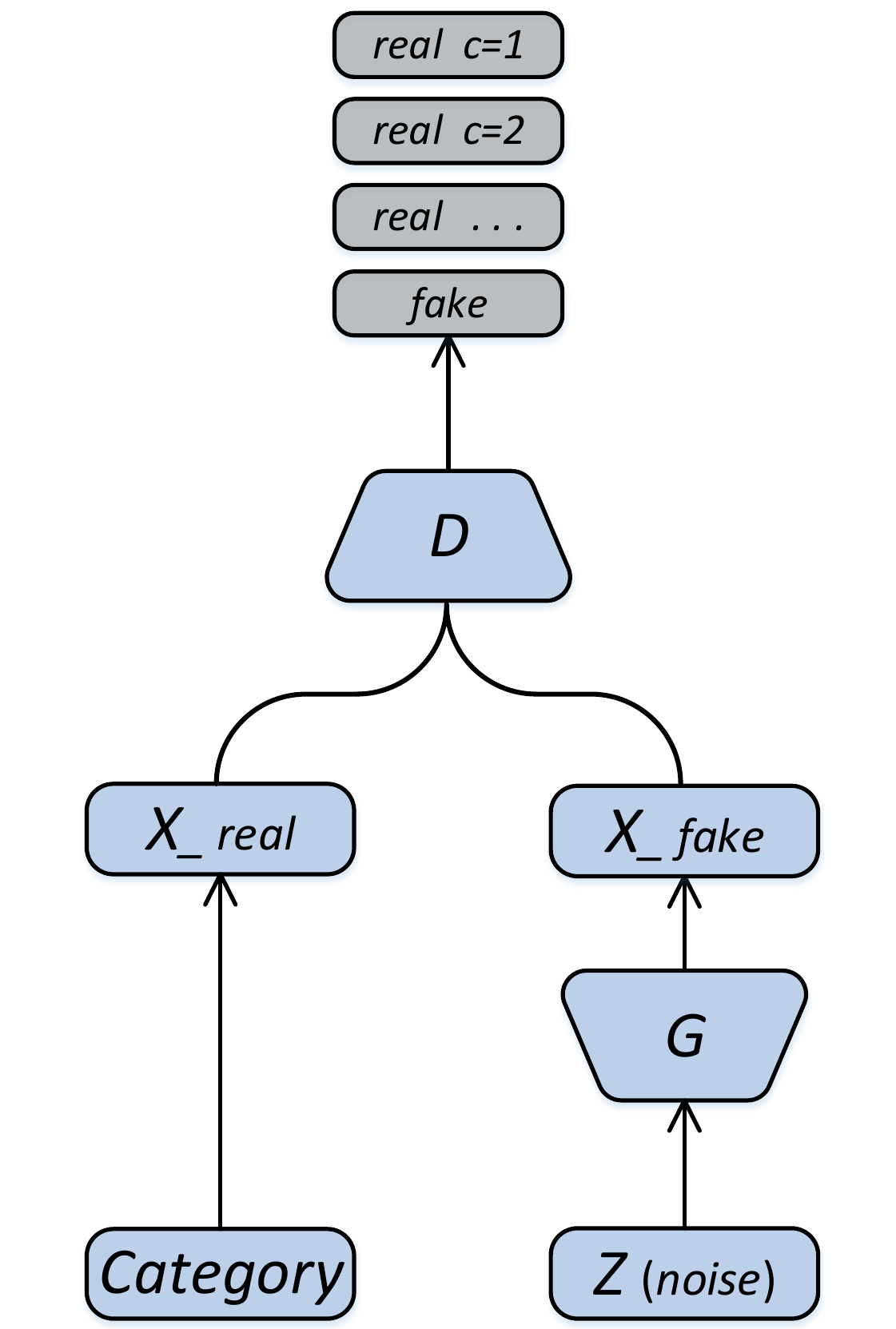}
        \caption{GAN-based attack}
        \label{fig:stru_uGAN}
    \end{subfigure}
    \begin{subfigure}[b]{.48\columnwidth}
        \includegraphics[width=\textwidth]{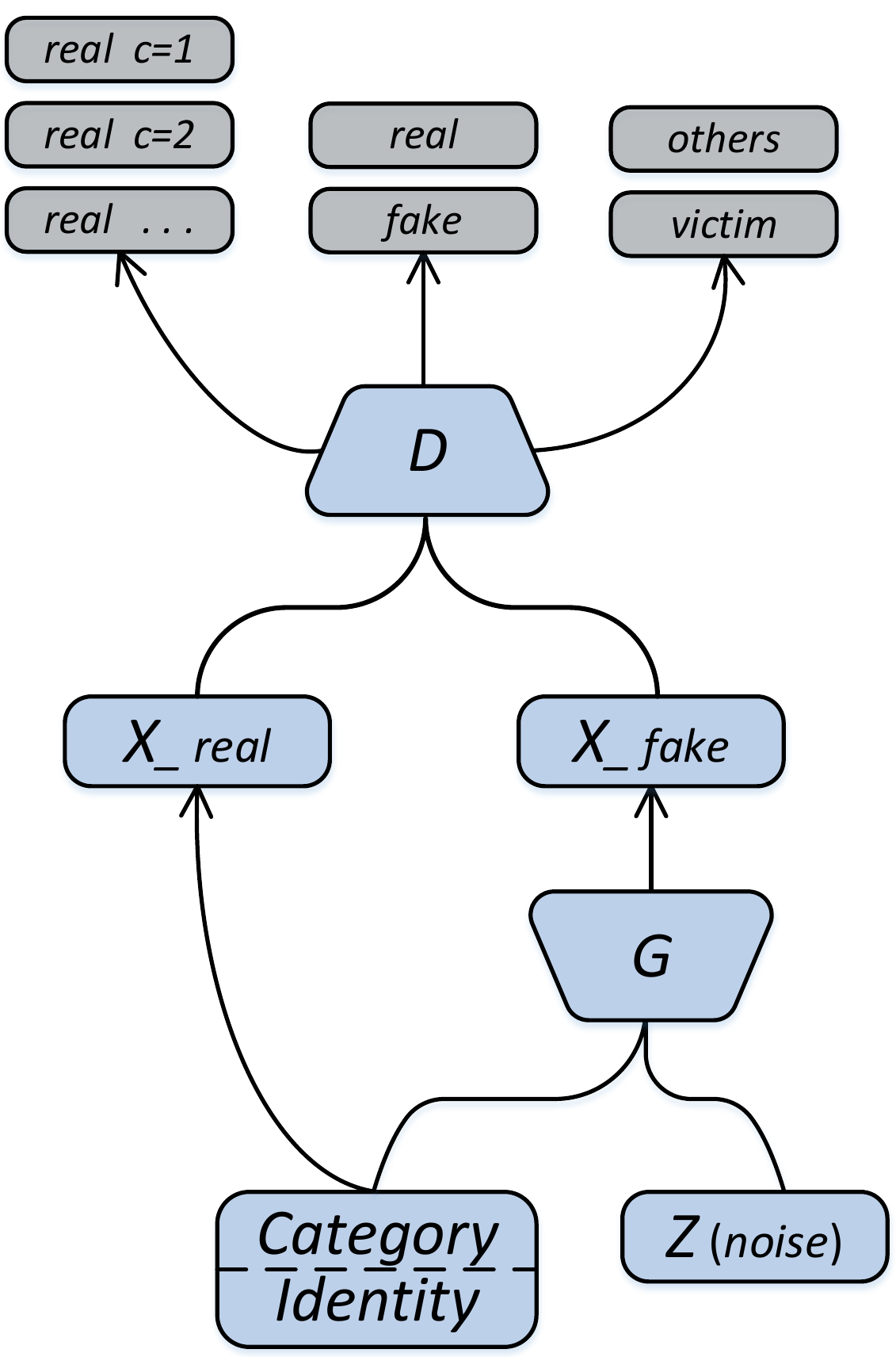}
        \caption{mGAN-AI}
        \label{fig:stru_mGAN-AI}
    \end{subfigure}
    \caption{Structure comparison between the GAN-based attack (a)~\cite{hitaj2017deep} and the proposed \alg~(b).}
    \label{fig:structures}
    \vspace{-2mm}
\end{figure}

\subsection{Structure of the Multi-task GAN}
\label{subsec:structure}
Specifically, the discriminator of \alg~is designed to achieve three tasks: 1) real-fake discrimination like standard GAN, 2) categorization on input samples, which performs like the shared model, further increasing the quality of generated samples~\cite{odena2016conditional}, and 3) identification on the input sample to distinguish the victim from other clients. The novel discrimination on identity is the key to violate
user-level privacy. Fig.~\ref{fig:stru_mGAN-AI} shows the multi-task GAN in the proposed \alg, where the discriminator $D$ shares the structure with the shared model except the output layer because the output format of the three tasks are different. Compared to the GAN-based attack~\cite{hitaj2017deep}, \alg~incorporates discrimination on identity that is also fed to $G$ as an important condition. The network structure of $D$ for each task in \alg~is illustrated in the following,
\begin{equation}
\begin{split}
D_{real} &= \text{Sigmoid}  \left( \text{FC}_{real} (L_{share}) \right) \\
D_{cat} &= \text{Softmax}  \left( \text{FC}_{cat} ( L_{share}) \right) \\
D_{id} &= \text{Sigmoid}  \left( \text{FC}_{id} (L_{share}) \right),
\end{split}
\label{eq:D_T}
\end{equation}
where $D_{real}$, $D_{id}$, and $D_{cat}$ denote the discriminator tasks of real-fake, identification, and categorization, respectively. Correspondingly, $\text{FC}_{id}$ and $\text{FC}_{cat}$ are the fully connection layer (i.e., output layer).
$L_{share}$ indicates the layers from the shared model except the output layer.

The generator $G$ accepts category and identity labels, as well as the noise (z), to conditionally yield user specified samples. The objectives are expressed in Eq.~\ref{eq:objectives}.
\begin{equation}
\begin{split}
\mathcal{L}_{real} = & \ \mathbb{E}_{x\sim p_{real}}[\log D_{real}(x)] + \\
~ & \ \mathbb{E}_{x\sim p_{fake}}[\log (1-D_{real}(x))]\\
\mathcal{L}_{cat} = & \
\mathbb{E}_{x,y\sim p_{fake}}[\text{CE}(D_{cat}(x), y)] \\
\mathcal{L}_{id} = & \ \mathbb{E}_{x\sim p_{victim}}[\log D_{id}(x)] + \\
~ & \ \mathbb{E}_{x\sim p_{other}}[\log (1-D_{id}(x))],\\
\end{split}
\label{eq:objectives}
\end{equation}
where $x \sim p_{victim}$ and $x \sim p_{other}$ denote the data sampled from the victim and the other clients, respectively. However, the samples from the victim or the other clients are inaccessible for the malicious server. Therefore, we estimate the representatives of each client based on their updates to the server. More details will be discussed later in section~\ref{subsec:client}. Note that $x\sim p_{real}$ indicates samples from an auxiliary dataset, instead of from the clients. Since the federated learning always requires a testing set to evaluate the learning process, and the testing set tends to consistent to the global distribution of all the clients, we could adopt such testing set as real data to implement the real-fake task. The $\text{CE}(\cdot)$ denotes the cross entropy. Updating $D$ will minimize $\mathcal{L}_{real} + \mathcal{L}_{id}$,
and updating $G$ will minimize $\mathcal{L}_{cat} - \mathcal{L}_{real} + \mathcal{L}_{id}$. The training process of \alg~will not affect the federated learning, thus referred to as passive attack whose training scheme will be detailed in the next section.

\subsection{Passive Attack}
\label{subsec:passive}
In the passive attack, the malicious server is assumed to be honest-but-curious, meaning that it only analyzes the updates from the clients, rather than modifying the shared model or introducing adversarial influence. The training scheme is briefed in Algorithm~\ref{alg:passive_attack}. The shared model $M$, discriminator $D$, and generator $G$ are initialized randomly. The inputs to \alg~are the shared model $M_t$ and updates $u_t^k$ ($k=1,2,\cdots,N$ and $N\ge 2$) from the $k$th client at each iteration $t$. Besides, an auxiliary set $X_{aux}$ is prepared for the training, and a victim indexed by $v$ is specified.

At the $t$th iteration, the current shared model $M_t$ is sent to the clients, then the clients send the updates back to the server. Following the rule of federated learning, those updates are averaged and added to $M_t$, obtaining the shared model for the next iteration. The discriminator $D$ is initialized by $M_t+u_t^v$ at each iteration to keep updated to the latest performance of categorization and to bias $D$ towards the victim, assisting the attack targeting the victim. As aforementioned that the identification task requires the samples/representatives from each client. The data representatives are calculated from those updates of the clients (for more details, please refer to section~\ref{subsec:client}). Finally, $D$ and $G$ are updated sequentially based on the objectives in Eq.~\ref{eq:objectives}, where the auxiliary set $X_{aux}$ is considered as the real data, and the generated samples $X_{fake}$ by $G$ are the fake data. The two datasets are fed to $\mathcal{L}_{real}$, achieving the real-fake task.
For the identification task (i.e., $\mathcal{L}_{id}$),
the representatives from $u_t^v$ is treated as samples from the victim, and those from $u_t^k~(k \ne v)$ are from others.

The update of $D$ and $G$ can be written as Eq.~\ref{eq:updateGAN}.
\begin{equation}
\begin{split}
D &= D - \eta_1\nabla_{\theta_D} (\mathcal{L}_{real} + \mathcal{L}_{id}) \\
G &= G - \eta_2\nabla_{\theta_G} (\mathcal{L}_{cat} - \mathcal{L}_{real} + \mathcal{L}_{id}),
\end{split}
\label{eq:updateGAN}
\end{equation}
where $\eta_1$ and $\eta_2$ denote the learning rates, and $\theta_D$ and $\theta_G$ indicate the parameters of $D$ and $G$, respectively.
\begin{algorithm}[t]
\caption{Passive attack of \alg}
\KwIn{The shared model $M_t$, updates $u^k_t$ from clients, auxiliary set $X_{aux}$, and target client $v$.}
\KwOut{The generator $G$ }
Initialize $M_0$, $G$, and $D$ \\
    \For{ t = 0 to T}
        {
            Send $M_{t}$ to the clients \\
            Receive updates from the clients, $\{u_t^1, u_t^2, \cdots, u_t^v, \cdots, u_t^{N} \}$\\
            Update the shared model using the updates, $\displaystyle M_{t+1} = M_{t} + \frac{1}{N}\sum_{k=1}^{N}u^k_t$\;
            Initialize $D$ by the shared model and update $u_t^v$, $D \leftarrow M_{t} + u_t^v$ \\
            \For{k=1 to N}
            {
              Calculate data representatives $X_k$ from $u_t^k$ \\
              \If{$k$ == $v$}
              {
                  label $X_k$ as victim
              }
              \Else
              {
                  label $X_k$ as others
              }
			}
            Get fake samples $X_{fake}$ from $G$ \\
            Update $D$ by minimizing
            $\mathcal{L}_{real} + \mathcal{L}_{id}$, feeding $X_{aux}$, $X_{fake}$, and $X_k$ ($k=1,2,\cdots,N$) \\
            Update $G$ by minimizing $\mathcal{L}_{cat} - \mathcal{L}_{real} + \mathcal{L}_{id}$
        }
\label{alg:passive_attack}
\end{algorithm}

To achieve better performance, a balanced training on $D$ is required, trained on balanced real and fake data. Since the parameters of $D$ are overwritten by the shared model which is trained more on the real data as the time of iteration increases, we propose to increase the training epoch of $D$ as $t$ increases, thus to compensate the lack of fake data during the training. The training will repeat until meeting the stop criteria: 1) the federated learning reaches the maximum iteration, or 2) the malicious server achieves desired results, i.e., the classification accuracy of the generated samples from $G$ converges above certain threshold as tested on the shared model.

\subsection{Active Attack}
\label{subsec:active}

The passive attack is invisible but needs to analyze all updates from the clients. A more efficient and powerful way of attacking a specified client would be to isolate the victim from the others, i.e., training \alg~on the victim alone by sending an isolated version of the shared model $M^{iso}$ to the victim. Intuitively, it could be considered that the malicious server contains an affiliated server that only connects to the victim, and the shared model between them is $M^{iso}$ instead of $M$. Initially, $M^{iso}_0 = M_0 = D$, where $M^{iso}$ and $D$ share the weights. Training \alg~on the affiliated server by following the similar training scheme to the passive attack, the generator would yield samples with higher quality and more distinguishable identity because \alg~is trained purely on the target real data. Note that the identification task (i.e., $\mathcal{L}_{id}$), as well as the calculation of representatives of clients, will be removed during the training because the victim is the only client w.r.t. the affiliated server.

Since the malicious server will ``actively'' send $M^{iso}$ to the victim to obtain specific information, we call it active attack. Although the active attack violates the original rule of federated learning, it does not introduce negative effect to the federated learning.

\subsection{Calculation of Client Representatives}
\label{subsec:client}
In the identification task, data from the victim and other clients are required to train the discriminator, while the sever cannot directly access those data. Aono et al.~\cite{aono2017privacy} recovered the client data based on the update from the client, which is accessible for a malicious server in the federated learning.
However, it is only suitable for a simple setting, where the shared model has to be a fully connected network, and the update is required to be obtained by training on a single sample.
Obviously, these limitations significantly impede the adaptation of \cite{aono2017privacy} to those widely adopted learning methods, e.g., convolution neural network, batch learning, etc.
Inspired by \cite{aono2017privacy}, we estimate the representative data of each client from its update sent to the server. The representatives of a client are defined to be synthesized samples that achieve the same update as the real samples after trained on the shared model.

As described in section \ref{subsec:passive}, $u^k_t$ is the update to the shared model $M_t$ after training the data of the $k$th client on $M_t$. Specifically, $u^k_t$ is obtained through backpropagation by minimizing the classification loss $\mathcal{L}$ on the shared model. By the same token, the corresponding representatives $X_k$ will be fed to $M_t$ and calculate the update $u^{X_k}_t$ through backpropagation by minimizing the same loss $\mathcal{L}$. Ideally, $u^k_t = u^{X_k}_t$. We adopt an optimization-based method to calculate $X_k$ based on the shared model, then the objective can be expressed as
\begin{equation}
\arg\underset{X_k}{\min}\| u_t^k - u^{X_k}_t \|_2,\; u^{X_k}_t = - \gamma \frac{\partial\mathcal{L}(X_k ; \theta_{M_t})}{\partial\theta_{M_t}},
\label{eq:rep_loss}
\end{equation}
where $\theta_{M_t}$ denotes the parameters of the share model $M_t$, and $\gamma$ is a scaling factor for balancing the magnitude of $u^{X_k}_t$. The initial representatives are drawn from random noise. 

Since optimization-based methods tend to introduce noise or artifacts, Eq.~\ref{eq:rep_loss} is further regularized by the total variation (TV)~\cite{mahendran2015understanding} as expressed in the following.
\begin{equation}
\mathcal{L}_{TV}(X_k) = \sum_{x\in X_k} \sum_{i,j}
\left( (x_{i,j+1}-x_{ij})^2+(x_{i+1,j} - x_{ij})^2 \right) ^ \beta,
\label{eq:rep_tv}
\end{equation}
where $X_k$ is a set of images, and $i$, $j$ are row and column indices of the image $x$.
The $\mathcal{L}_{TV}$ computes the neighborhood distance to encourage spatial smoothness of an image.

Finally, the objective of achieving valid representatives $X_k$ for the $k$th client is
\begin{equation}
\arg\underset{X_k}{\min}\left\| u_t^k - u^{X_k}_t \right\|_2 + \lambda \mathcal{L}_{TV}(X_k),
\label{eq:loss_total}
\end{equation}
where $\lambda$ balances the effect of TV. We calculate $X_k$ by using the box-constrained L-BFGS.
Empirically, valid $X_t$ could be achieved after several updates. The experiments in section~\ref{sec:experiment} validate the effectiveness of the representatives in the identification task.

\section{Experimental Evaluation}
\label{sec:experiment}
In this section, we first clarify the dataset and experiment setup in Sec.~\ref{subsec:datasets}. Then, effectiveness of the proposed \alg~is validated in Sec.~\ref{subsec:GANAI_Evaluation}. Finally, section~\ref{subsec:Comparison} conducts qualitative and quantitative comparison between \alg~and another two typical attacks, i.e., the model inversion (MI) attack and GAN-based attack.

\subsection{Datasets and Experiment Setup}
\label{subsec:datasets}
\subsubsection{MNIST}
It contains 70,000 handwritten digits images from 0 to 9 (i.e., 10 classes). The images are in gray scale with the size of 28$\times$28, and they are divided into the training set (60,000 samples) and testing set (10,000 samples).

\subsubsection{AT\&T}
The AT\&T dataset consists of facial images from 40 different persons, namely 10 images per person. The images are in gray scale with the size of 92$\times$112. They were taken at different times and with large variation in facial expression (smiling or w/o smiling) and facial details (glasses or w/o glasses). We resize the images into 64$\times$64.


\subsubsection{Experiment setup}
As discussed in Sec.~\ref{sec:attack}, the proposed \alg~involves three components: 1) classifier (i.e., the shared model), 2) discriminator, and 3) generator. The classifier and discriminator are constructed by convolutional neural networks, and they share the structure except the output layer. The generator is constructed with deconvolution layers.
Tables~\ref{tab:nets_MINIST} and \ref{tab:nets_att} show the network structures for MNIST and AT\&T, respectively.
For the generator, the kernel size is $5\time 5$ with the stride of 2. The input to the generator is formated as concatenation of random noise, categorical label, and identity label (Fig.~\ref{fig:stru_mGAN-AI}), whose length are 100, 1, and 1, respectively. The embedding layer squeezes the length of input to 100.
For the classifier and discriminator, the kernel size is $3\times 3$. Their output has the similar format as the input of the generator, i.e., the first digits for reality, the next $C$ digits for category, and the last digit for identity. The $C$ indicates the number of classes for different datasets, i.e., $C=10$ for MNIST and $C=40$ for AT\&T.
The activation functions ReLU and LReLU are adopted in the generator and classifier/discriminator, respectively.
In addition, batch normalization is used at the intermediate layers of the generator.

In the training of the classifier (i.e., shared model), the SGD optimizer is adopted with the learning rate of 0.002. For discriminator and generator (i.e., multi-task GAN), the Adam optimizer is used with the learning rate and momentum term to be 0.0002 and 0.5, respectively. In calculating the representatives, the box-constrained L-BFGS is employed, where $\lambda = 0.00015$, $\beta=1.25$. 

\begin{table}[t]
	\centering
	\caption{Network structure for MNIST}
	\begin{tabular}{c|l}
		\hline
		\begin{tabular}{@{}c@{}}Classifier/\\Discriminator
		\end{tabular} &
		\begin{tabular}{@{}l@{}}
			$28^2\times1 \xrightarrow{\text{Conv (stride = 2)}} 14^2\times32 \xrightarrow{\text{Conv (stride = 1)}}  $\\
            $14^2\times64 \xrightarrow{\text{Conv (stride = 2)}} 7^2\times128 \xrightarrow{\text{Conv (stride = 1)}} $\\
            $7^2\times256 \xrightarrow{\text{FC}} 12,544\xrightarrow{\text{FC, Softmax}} (1, 10, 1)$
		\end{tabular}\\
		\hline
		Generator &
		\begin{tabular}{@{}l@{}}
			$(100,1,1)\xrightarrow{\text{Embedding}} 100 \xrightarrow{\text{FC}} 3^2\times384 \xrightarrow{\text{Deconv}} $\\
            $ 7^2\times192 \xrightarrow{\text{Deconv}} 14^2\times96  \xrightarrow{\text{Deconv, tanh}} 28^2\times1$
		\end{tabular}
		\\
		\hline
	\end{tabular}
	
	\begin{tablenotes}
		\footnotesize
		\item 
	\end{tablenotes}
	\label{tab:nets_MINIST}
    \vspace{-2mm}
\end{table}

\begin{table}[t]
	\centering
	\caption{Network structure for AT\&T}
	\begin{tabular}{c|l}
		\hline
		\begin{tabular}{@{}c@{}}Classifier/\\Discriminator
		\end{tabular} &
		\begin{tabular}{@{}l@{}}
			$64^2\times1 \xrightarrow{\text{Conv (stride = 2)}} 32^2\times32 \xrightarrow{\text{Conv (stride = 1)}}  $\\
            $32^2\times64 \xrightarrow{\text{Conv (stride = 2)}} 16^2\times128 \xrightarrow{\text{Conv (stride = 2)}}  $\\
            $8^2\times256 \xrightarrow{\text{FC}} 16,384\xrightarrow{\text{FC, Softmax}} (1,40,1)$
		\end{tabular}\\
		\hline
		Generator &
		\begin{tabular}{@{}l@{}}
			$(100,1,1)\xrightarrow{\text{Embedding}}
            100 \xrightarrow{\text{FC}}4^4\times512\xrightarrow
            {\text{Deconv}} $\\
            $8^2\times256 \xrightarrow{\text{Deconv}}  16^2\times128 \xrightarrow{\text{Deconv}} $\\
            $32^2\times256 \xrightarrow{\text{Deconv}}  64^2\times256 \xrightarrow{\text{Deconv, tanh}} 64^2\times1$
		\end{tabular}
		\\
		\hline
	\end{tabular}
	
	\begin{tablenotes}
		\footnotesize
		\item $m^2\times n$ denotes the size of a layer, i.e., $m\times m$ map with $n$ channels.
	\end{tablenotes}
	\label{tab:nets_att}
    \vspace{-2mm}
\end{table}

\subsection{Client-level Privacy Attack by \alg}
\label{subsec:GANAI_Evaluation}
This section evaluates the effectiveness of passive \alg~by comparing the reconstructed samples with the real samples of the victim. Instead of evenly splitting the training data to the clients, we simulate biased data distribution on clients according to the setting of federated learning. In addition, opposite to \cite{hitaj2017deep} that assumed each client consists of a single class, we achieve more flexible condition that each client could own samples from multiple classes. Note that different clients may share the samples from the same class.


\begin{figure}[ht]
	\centering
    \rotatebox{90}{
        \footnotesize Output \hspace{.05cm} \footnotesize $X_v$ \hspace{.1cm} \footnotesize $X_k$ \hspace{.3cm}\footnotesize Victim \hspace{.001cm} \footnotesize Other
    }
	\includegraphics[width=.95\columnwidth]{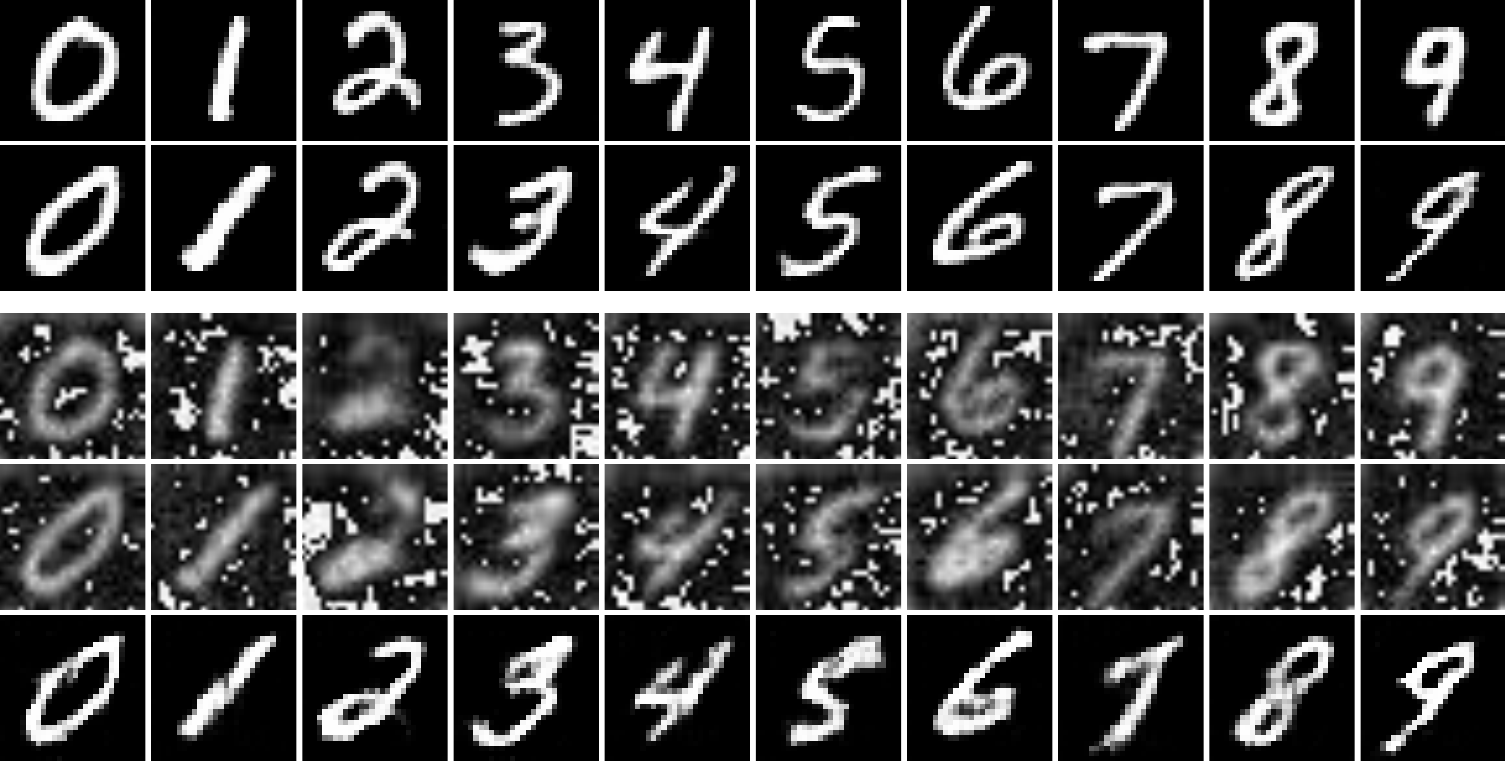}
	\caption{The results of \alg~on the MNIST dataset. The first two rows are the real samples from the other clients and the victim. The next two rows are corresponding representatives calculated from other ($X_k$, $k \ne v$) and victim ($X_v$).
The last row is the reconstructed samples by \alg, which are similar to those of the victim, presenting larger rotation as compared to those of the other clients.}
	\label{fig:mnist}
\end{figure}

\subsubsection{Evaluation on the MNIST dataset} We set the number of client $N=10$.
Each client randomly draws 100 samples from three random classes as its private data.
To gain distinguishable properties for the victim, we rotate the samples of the victim by InfoGAN~\cite{chen2016infogan}, which could synthesize rotated version of the digits. Ideally, \alg~should reconstruct rotated digits owned by the victim.
Fig.~\ref{fig:mnist} shows the reconstructed client-level samples by \alg~in the federated learning.
The first two rows are the samples from other clients and the victim, respectively. The 3rd and 4th rows are the corresponding representatives ($X_k$ in Algorithm~\ref{alg:passive_attack}) calculated from other clients and the victim. The last row shows the reconstructed samples, which is indistinguishable from the second row (victim) and different from the first row (other) in rotation obviously.
Note that ten classes are presented in Fig.~\ref{fig:mnist} for the victim because the experiment is repeated, so that the victim could walk through all possible classes.
This demonstrates that \alg~can successfully and precisely recover the provide data of the victim, i.e., compromising the client-level privacy.

\begin{figure}[t]
	\centering
	\includegraphics[width=\columnwidth, trim=0 25 0 0, clip]{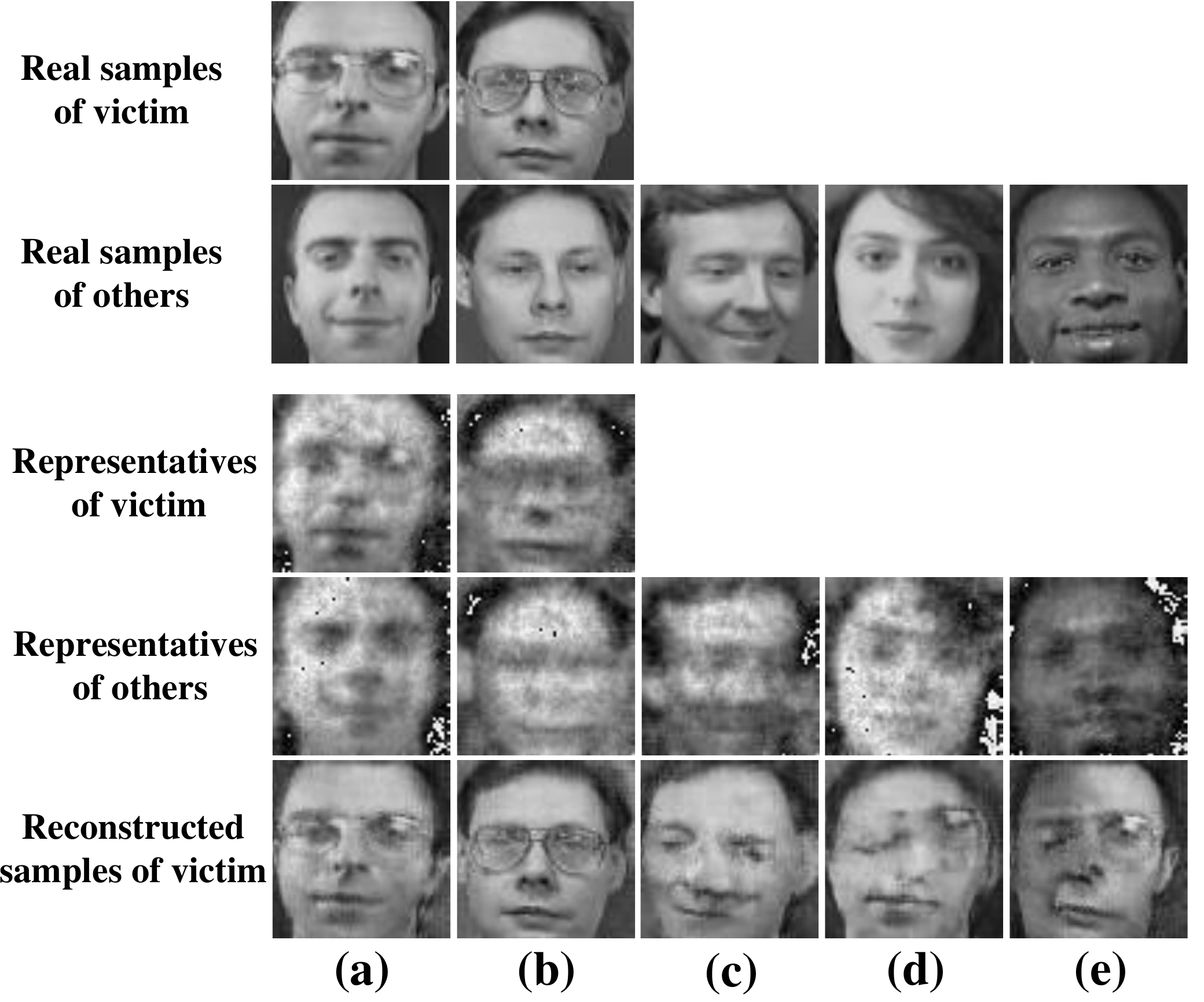}
	\caption{The results of \alg~on the AT\&T dataset. The top two rows are real samples from the victim and other clients, respectively. The corresponding representatives are shown in the next two rows. The last row is the reconstructed samples, which show that \alg~would specifically recover client-dependent property, i.e., wearing glasses.}
	\label{fig:att}
	\vspace{-4mm}
\end{figure}

\subsubsection{Evaluation on the AT\&T dataset}
The similar experiment is performed on the AT\&T dataset, where the number of clients $N=10$.
The distinguishable property of the victim is assigned to be wearing glasses. As shown in Fig.~\ref{fig:att}, the first row shows the private data of the victim, i.e., two faces wearing glasses. The second row are faces from other clients that do not wear glasses. Note that the first two faces in the second row are the same person as in the first row. Therefore, successfully attack to  the victim should be reconstruction of the first two person wearing glasses. The 3rd and 4th rows are corresponding representatives. The last row are reconstructed samples, where the first two samples are identical to the victim while the rest are significantly distorted. This demonstrates that \alg~could only clearly reconstruct the private data of the victim, i.e., \alg~specifically attacks the victim.

Note that \alg~does not violate the differential privacy (DP), which aims to prevent the recovery of specific samples used during the training. In other words, DP tries to make the adversary fail to tell whether a given sample belongs to the training set.
However, without inferring the membership of a given sample, it can still generate samples distinguishable from real samples as demonstrated in Figs. \ref{fig:mnist} and \ref{fig:att}, which obviously leads severe privacy violation.
Besides, convergence of the shared model requires a relatively loose privacy budget when applying DP in the federated learning~\cite{melis2018inference}, which means the magnitude of the perturbation should be carefully controlled to ensure high-utility updates used for training the multi-task GAN in \alg.

\begin{figure}[t]
	\centering
	\includegraphics[width=0.95\columnwidth]{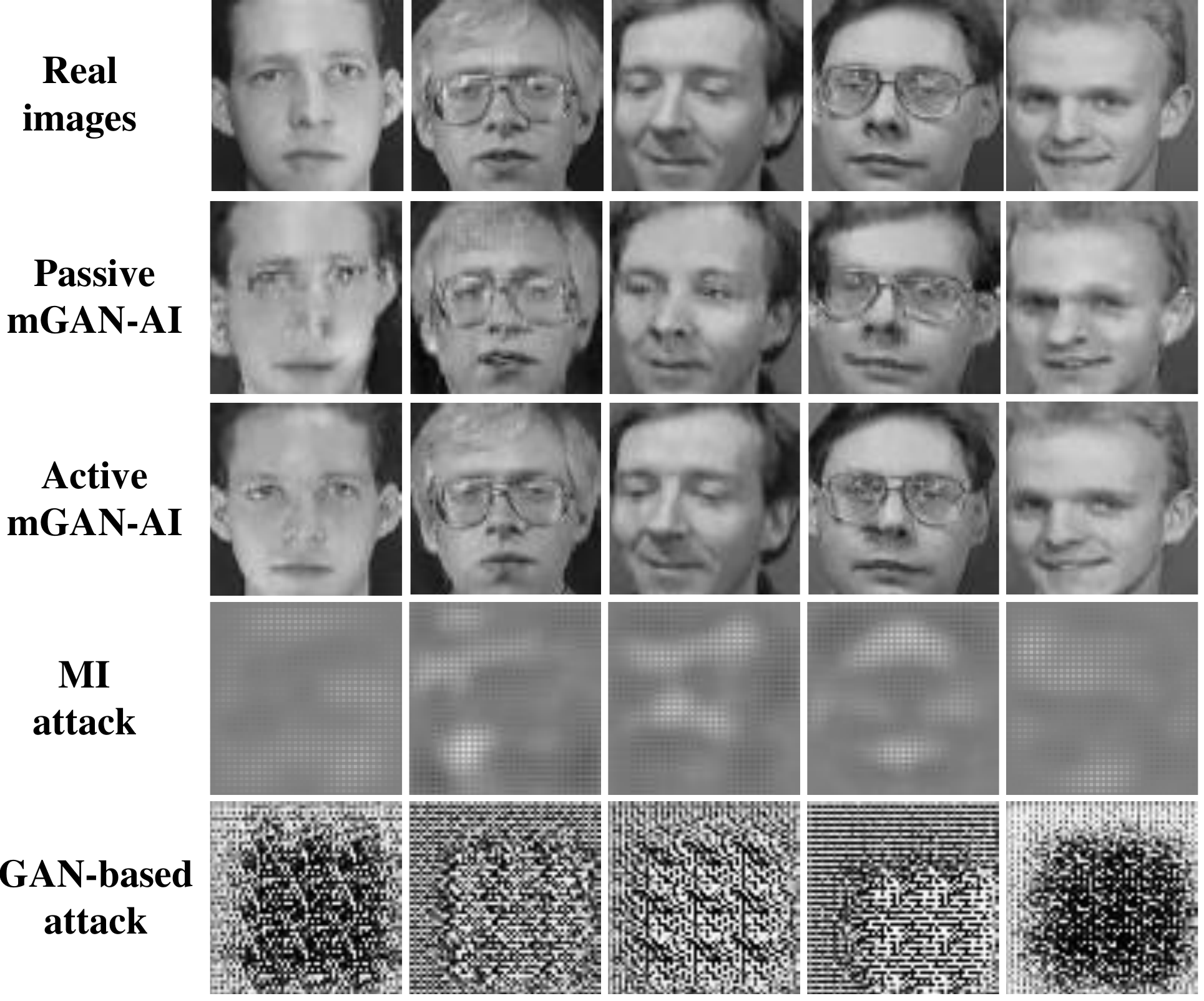}
	\caption{Comparison of reconstructed samples by \alg, MI attack, and GAN-based attack. The top row is the real samples, and the rest are the reconstructed samples by the passive \alg, active \alg, MI attack, and GAN-based attack, respectively.}
	\label{fig:compared}
	\vspace{-2mm}
\end{figure}

\subsection{Quantitative and Qualitative Comparison}
\label{subsec:Comparison}
This section compares \alg~with two state-of-the-art attacking models, i.e., inversion attack (MI) and GAN-based attack, in aspects of reconstruction quality of the victim and side effect to the federated learning.
For fair comparison, the same experiment setup is followed in the federated learning, the same dataset AT\&T is adopted, and the number of clients $N=20$.
The results are shown in Fig.~\ref{fig:compared}, where the first row are real samples from the victim. The second and third rows are passive attack and active attack of \alg, respectively. The last two rows are reconstructions by MI attack and GAN-based attack, respectively.

Comparing the results from the proposed \alg~with MI and GAN-based attack, \alg~generates samples with much higher quality and more identical to the real samples.
The results of MI attack is significantly blurry, consistent with the results in \cite{hitaj2017deep,shokri2017membership}.
The failure is mainly caused by simply maximizing the model output w.r.t. a target label, which would lead to uninterpretable samples when dealing with complicated neural networks.
sharing the spirit with another threat for deep learning --- adversarial examples~\cite{szegedy2013intriguing}.
The results of GAN-based attack do not converge at all because it heavily relies on the introduced ``adversarial influence'' which would be averaged (become trivial) before adding to the shared model. The results of GAN-based attack are consistent with those in its original work, where unrecognizable images were generated without the adversarial influence.
Comparing the passive and active attack of \alg, the active attack performs better but it would slightly degrade the learned model in the federated learning. By contrast, the passive learning does not effect the learning model, which will be demonstrated later.

To quantitatively evaluate the generated images, the inception score~\cite{salimans2016improved} is used to statistically compare the three attack models. Inception score has been widely adopted in image quality evaluation, especially in the area of image synthesis.
We generate 400 samples from each of the three attacks and compute the inception score for each method. The results are shown in Table~\ref{tab:inception_score}, where the proposed \alg~shows higher score than MI and GAN-based attack, demonstrating that \alg~generates samples with higher quality.
\begin{table}[t]
\centering
\caption{Quantitative Comparison on Inception Score}
\begin{tabular}{C{3cm}|C{3cm}}
\hline
  &   Inception Score \\ \hline
  Real Images &  1.55 $\pm$ 0.04  \\
  Passive \alg &  1.42 $\pm$ 0.02  \\
  Active \alg &  1.61 $\pm$ 0.05 \\
  GAN-based attack &  1.18 $\pm$ 0.03  \\
  MI attack &  1.01 $\pm$ 0.03  \\
\hline
\end{tabular}
\label{tab:inception_score}
\vspace{-4mm}
\end{table}

\begin{figure}[ht]
	\centering
	\includegraphics[width=0.95\columnwidth]{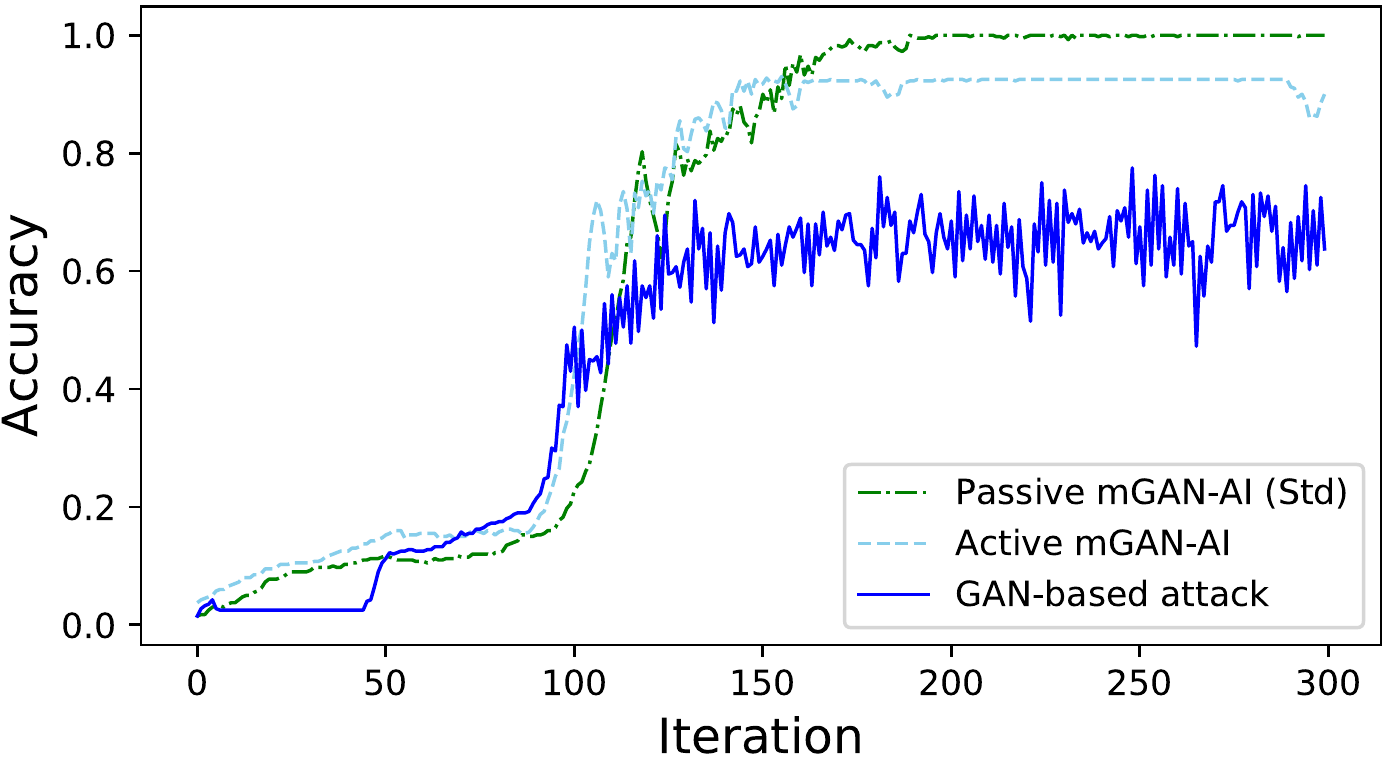}
	\caption{The accuracy of the shared model at each iteration in the federated learning when mounting different attacks.}
	\label{fig:side_effect}
	\vspace{-2mm}
\end{figure}

Finally, we investigate the side effect of the three attacking models to the federated learning.
Fig.~\ref{fig:side_effect} shows the accuracy of the shared model at each iteration in the federated learning when performing different attacks, i.e, passive \alg, active \alg, and GAN-based attack. The GAN-based attack significantly reduces the accuracy and presents drastic oscillation (non convergence) because it introduces adversarial influence. The passive \alg~achieves the highest accuracy because it does not affect the training process of federated learning. The active \alg~gets a bit lower but stable accuracy. With this cost, however, it achieves better reconstruction quality as shown in Fig.~\ref{fig:compared}.

\section{Conclusions}
\label{sec:conclusion}

This paper investigated the privacy risk of federated learning, which is considered as a privacy-preserving learning framework working in a decentralized manner. Against federated learning, we proposed a generic and practical reconstruction attack named \alg, which enables a malicious server to not only reconstruct the actual training samples, but also target a specific client and compromise the client-level privacy. The proposed attack does not affect the standard training process, showing obvious advantages over the current attack mechanisms.
The extensive experimental results on two benchmark datasets demonstrate that the \alg~can reconstruct samples that resemble the victim's training samples, outperforming the state-of-the-art attacking algorithms.

\bibliographystyle{IEEEtran}
\balance
\bibliography{DP}
\end{document}